\documentclass{article}
\usepackage[uppertitlebar]{ishwaran}
\RequirePackage[authoryear]{natbib}
\bibpunct{[}{]}{,}{n}{}{;}

\usepackage{enumitem}
\usepackage{authblk}
\usepackage{tabularx}
\usepackage{makecell}
\usepackage{capt-of}

\usepackage[linesnumbered,ruled,vlined,nosemicolon]{algorithm2e}
\usepackage{hyperref}
\SetKwFor{ForEach}{for each}{do}{end}

\def\pp{\mathcal{P}}
\def\nn{\mathcal{N}}
\def\rr{\mathcal{R}}
\def\PP{\mathbb{P}}
\def\RR{\mathbb{R}}
\def\z{\zeta}

\def\warp{\texttt{warp}}
\def\joint{\texttt{joint}}
\def\support{\texttt{support}}

\def\varPro{\texttt{varPro}}

\graphicspath{{figures/}}

\title{General OOD Detection via Model-aware \\ and Subspace-aware Variable Priority}

\author{
  \href{https://orcid.org/0000-0002-1386-1315}{\includegraphics[scale=0.16]{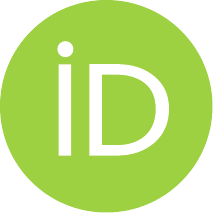}Min Lu\textsuperscript{*}}
  \and
  \href{https://orcid.org/0000-0003-2758-9647}{\includegraphics[scale=0.16]{orcid.pdf}Hemant Ishwaran}
  \thanks{Division of Biostatistics, Miller School of Medicine, University of Miami, Miami, USA}
  \thanks{Corresponding Author\\
    Min Lu (\href{mailto:m.lu6@umiami.edu}{m.lu6@umiami.edu}) \\
    Hemant Ishwaran (\href{mailto:hishwaran@med.miami.edu}{hishwaran@med.miami.edu})}
}

\date{}

\renewcommand{\runningtitle}{General OOD Detection via Model and Subspace Awareness}

\begin{document}
\maketitle

\begin{abstract}

Out-of-distribution (OOD) detection flags test inputs that depart from
the data used to train a model. For structured tabular problems with
regression or survival outcomes, existing methods remain limited
because many OOD detectors are designed for classification or
unstructured data. We introduce a tree based method that uses the rule
structure of a supervised forest to determine the predictive subspace,
the model-aware reference neighborhood, and the final OOD score in a
single construction. The score compares each test input with forest
selected reference cases only on prediction relevant variables,
reducing signal dilution from nuisance coordinates. Across synthetic
and real data benchmarks, the method performs especially well for
subtle targeted feature shifts and changes in dependence. An
esophageal cancer survival study further shows how OOD scores can
reveal lymphadenectomy related shifts relevant to surgical guidelines.

\end{abstract}


\section{Introduction}\label{sec1}

Most prediction models are designed to return predictions, not to
indicate when a new case is poorly supported by the data.  A new input
may differ from the training population in a single influential
covariate, in a subtle combination of variables, or in the dependence
structure among covariates, yet the model may still return a numerical
prediction, survival curve, or risk score that may appear routine even
though the test case should be considered suspicious.  In settings
such as clinical medicine and medical informatics, where predictions
may influence diagnosis, monitoring, or treatment, this can have
serious consequences.  \emph{Out-of-distribution} (OOD) detection
addresses this problem by flagging inputs that are poorly represented
by the data used to train the model
\citep{hendrycks2016baseline,salehi2022unified}. Because the forms of
distributional departure that arise after deployment are rarely known
in advance, an OOD detector must often be learned from
\emph{in-distribution} (ID) data alone.

We study OOD detection for supervised tabular prediction with
continuous or time-to-event outcomes. Much of the OOD literature has
been developed for image and text classification, where class
probabilities, logits, and learned representations provide natural
ingredients for OOD scoring. Regression and survival problems lack this
class-based structure. When such problems are recorded in tabular form,
however, the covariates are explicit, and the trained model can reveal
how those covariates support prediction. The key idea in this paper is
to use this model-derived structure to assess unusualness relative to
the prediction task rather than relative to the covariate distribution
alone. Specifically, we use the rule structure of a supervised random
forest to identify coordinates that contribute to prediction and to
isolate reference training cases that are locally relevant for a test
input. The resulting OOD score compares the test input with those
reference cases within the prediction-relevant covariate subspace.

To place this approach in the standard distribution-shift framework,
write $\PP_{X,Y}$ for the joint law of $(X,Y)$ in the training
population, with marginal law $\PP_X$ and conditional law
$\PP_{Y\mid X}$. Let $\PP^*_{X,Y}$ denote the corresponding test law,
with marginal $\PP^*_X$ and conditional law $\PP^*_{Y\mid X}$. In the
tabular settings considered here,
$X=(X^{(1)},\ldots,X^{(d)})\in\RR^d$ is the structured covariate vector
used to predict the outcome $Y$. The test distribution is OOD relative
to the training distribution whenever
$\PP^*_{X,Y}\neq\PP_{X,Y}$. Such a difference may arise through
covariate shift, in which $\PP_X$ changes while $\PP_{Y\mid X}$ remains
fixed; conditional or concept shift, in which $\PP_{Y\mid X}$ changes;
or changes in both components
\citep{quionero2009dataset,storkey2009training,
shimodaira2000covariate,sugiyama2007covariate,
morenotorres2012unifying,bendavid2010theory}.

This taxonomy describes how training and test laws can differ, but it
does not by itself determine which differences should make an
individual input suspicious for a supervised predictor. A change in
$\PP_X$ along nonpredictive coordinates can make a case atypical in the
full covariate space while leaving it well supported for prediction
\citep{wu2024low}. Conversely, a localized change involving a small
number of influential variables, or a change in their dependence
structure, may be difficult to recognize in the full covariate space
but important for prediction. A supervised OOD score should therefore
reflect how variables enter the prediction rule rather than treating
every detectable departure in $\PP_X$ as equally consequential
\citep{rabanser2019failing}. Our score implements this principle by
comparing each input with forest-selected training cases in the
covariate subspace used for prediction.

Although this approach is very general, we note a limitation of our method,
shared by other OOD input methods, is that it cannot detect shifts
that leave the covariate law unchanged. For a new case, our score uses
only the case covariates and the fixed forest learned from the ID
covariates and outcomes; no outcomes from the test distribution are
observed when the score is computed. Consequently, a pure concept
shift, with $\PP^*_{Y\mid X}\neq\PP_{Y\mid X}$ but $\PP^*_X=\PP_X$, is
OOD under the joint law but leaves no observable covariate signal for
the detector. Detecting such a shift requires outcomes from the test
distribution or additional assumptions, and is outside the scope of
this paper.

\subsection{Predictive subspace}

We formalize the part of the covariate space relevant to prediction
using a predictive subspace. Let
$S_0 \subset \{1,\ldots,d\}$ denote a set of coordinates through which
the training conditional law factors. For any $x\in\RR^d$, write
$x^{(S_0)}=(x^{(j)})_{j\in S_0}$. We suppose
$$
\PP(Y\in A\mid X=x)
=
\PP(Y\in A\mid X^{(S_0)}=x^{(S_0)})
$$
for every measurable set $A$ and $\PP_X$-almost every $x$. Thus, under
the training distribution, the conditional law of $Y$ depends on $X$
only through $X^{(S_0)}$. We do not require $S_0$ to be unique or
minimal. Any set satisfying this condition provides a valid population
target, although the purpose of the restriction is to remove
coordinates that are unnecessary for prediction.

The set $S_0$ describes the population target of the subspace
restriction, but it is unknown in practice. Our method uses variable
priority from a supervised forest to construct an empirical signal set
$\widehat{S}$ and then uses the same forest to determine the local
reference set for a test input. Thus, $S_0$ motivates the subspace
restriction, whereas $\widehat{S}$ is the learned subspace used by the
score.

\subsection{Approach}

The proposed method rests on two properties. First, the OOD score
depends on the trained supervised model, not only on the geometry or
density of the covariates. Second, once the relevant training cases are
identified, the final comparison uses only coordinates containing
predictive information.

The terms \emph{model-aware} and \emph{subspace-aware} refer to these
two operational properties. Fix a training procedure $\pp$ and hold any
internal algorithmic randomness fixed. Given training data
$\mathcal{L} = \{(x_i, y_i)\}_{i=1}^n$, write
$\hat{f} = \pp(\mathcal{L})$ and $x_{1:n} = (x_1, \ldots, x_n)$.

\begin{definition}[Model-aware OOD score]\label{def:modelaware}
An OOD score $d(x^*; x_{1:n}, \hat{f})$ is \emph{model-aware} if the
trained supervised model $\hat{f}$ enters the construction of the
score. Thus, holding $x^*$ and $x_{1:n}$ fixed, changing the training
outcomes can change the score through $\hat{f}$.
\end{definition}

\begin{definition}[Subspace-aware OOD score]\label{def:subspaceaware}
Let $S \subset \{1,\ldots,d\}$ be a selected set of coordinates, and
let $\mathcal{R}(x^*; x_{1:n}, \hat{f}) \subseteq \{x_1,\ldots,x_n\}$
denote the reference set used for test input $x^*$. An OOD score is
\emph{subspace-aware with respect to $S$} if, after the reference set
has been formed, the final discrepancy depends on $x^*$ and the
reference points only through their coordinates in $S$.
\end{definition}

Our construction satisfies both properties with a single learned
object. We train a supervised forest on the ID data and use its
decision rules in two connected ways. First, the rules identify the
empirical signal set $\widehat{S}$. Second, for a test point $x^*$, we
relax the rules containing $x^*$ one selected coordinate at a time,
rank the training points, and form a neighborhood $\nn_K(x^*)$. This
neighborhood is the reference set
$\mathcal{R}(x^*; x_{1:n}, \hat{f})$ in
Definition~\ref{def:subspaceaware}. The final OOD score averages the
distance from $x^*$ to this reference set, computed only on
$\widehat{S}$,
$$
d(x^*) = \frac{1}{K} \sum_{z \in \nn_K(x^*)}
D_{\widehat{S}}(x^*, z),
$$
where $D_{\widehat{S}}$ denotes a distance using only the coordinates
in $\widehat{S}$, with optional weights reflecting their relative
importance. Thus, the score depends on $\hat{f}$ through both
$\widehat{S}$ and $\nn_K(x^*)$, satisfying
Definition~\ref{def:modelaware}, while its final discrepancy uses only
$\widehat{S}$, satisfying Definition~\ref{def:subspaceaware}.

\subsection{Main Contribution and Evaluation}

Our main contribution is an integrated model-aware and subspace-aware
OOD score for tabular supervised learning. A single trained supervised
forest identifies prediction-relevant variables, constructs a local
reference set for each test input through learned rule regions, and
computes a weighted distance on the selected variables. This
construction differs from approaches that first reduce dimension or
select features and then apply a standard distance, nearest-neighbor,
or density-based anomaly score.

The empirical sections evaluate the construction in three ways. First,
matched ablations separate the effects of variable priority selection,
the forest rule neighborhood, and variable weighting. Second, benchmark
experiments compare the method with existing OOD detectors across
synthetic, regression, and high-dimensional survival settings. Third,
an esophageal cancer case study shows how the scores inform a clinical
survival analysis. Software implementing the method is available in the
CRAN package \varPro. Public benchmark datasets and microarray data
used in the experiments are available from the cited references.

\subsection{Organization}

The paper is organized as follows. Section~\ref{sec2} reviews related
work and explains how our approach differs. Section~\ref{sec3}
develops the methodology, and Section~\ref{sec4} describes the
comparison methods. Section~\ref{sec5} presents the benchmarking
results. Section~\ref{sec6} applies the method to a survival analysis
of esophageal cancer. Section~\ref{sec7} summarizes and discusses
limitations.

\section{Related Work}\label{sec2}

Research on out-of-distribution (OOD) behavior spans anomaly
detection, open-set recognition, selective prediction, domain
generalization, and OOD detection
\citep{salehi2022unified,yang2024generalized,tamang2025handling,
lu2025out}. Recent surveys distinguish methods developed for
supervised learning
\citep{salehi2022unified,yang2024generalized,tamang2025handling}
from methods tailored to structured domains such as graphs and time
series \citep{zhang2024graphshift,li2025graphood,wu2025tsood}. Our
focus is test time OOD scoring for supervised tabular models,
especially regression and survival analysis, where the score should
use what the fitted model has learned about prediction relevant
covariates.

\subsection{Dataset shift and prediction relevance}

Dataset shift is commonly described as a difference between the joint
training and test distributions. Standard taxonomies distinguish
changes in the covariate distribution from changes in the conditional
law of the outcome
\citep{quionero2009dataset,storkey2009training,
morenotorres2012unifying}. Covariate shift refers to a change in
$\PP_X$ with $\PP_{Y\mid X}$ stable, while concept shift refers to a
change in the conditional relationship between predictors and outcome
\citep{shimodaira2000covariate,sugiyama2007covariate,
morenotorres2012unifying}. Domain adaptation theory gives related
bounds for learning when training and test distributions differ
\citep{bendavid2010theory}.

In supervised tabular prediction, a covariate departure is most
relevant when it occurs in directions that affect the fitted
prediction rule. Departures confined to nuisance coordinates may make
a point unusual under $\PP_X$ without making it unusual for the
prediction task. This motivates an OOD score that uses the fitted
model to determine both the variables and the local training cases
used in the comparison.

\subsection{Model based OOD scores}

Most OOD detection methods were developed for classification, where
class probabilities, logits, and learned representations provide
natural scoring rules. Examples include maximum softmax probability
\citep{hendrycks2016baseline}, ODIN \citep{liang2018odin},
energy-based scores \citep{liu2020energy}, Mahalanobis scores in
learned feature space \citep{lee2018mahalanobis}, deep nearest
neighbors \citep{sun2022deepknn}, and posterior or representation
shaping methods
\citep{ming2022poem,zhang2024shapeood,chen2024sparsity}. Related
subspace and spectral methods decompose or project learned
activations before scoring
\citep{zongur2025actsub,zhu2024projood,fang2024kernelpca}. These
approaches can be highly effective for image and text benchmarks, but
they are closely tied to discrete labels, deep representations, or
global feature spaces, and are not directly designed for continuous
outcomes or survival data.

For regression and related continuous outcome problems, OOD detection
has often followed an uncertainty based paradigm. Deep ensembles
\citep{Lakshminarayanan2017DeepEnsembles}, Monte Carlo dropout
\citep{Gal2016DropoutBayesian}, evidential regression
\citep{Amini2020EvidentialRegression}, prior networks
\citep{Malinin2019PriorNetworks}, predictor entropy methods
\citep{Pequignot2020RegressionOOD}, Gaussian processes, and deep
kernel learning
\citep{laGP,wilson2016dkl,vanamersfoort2021dklcollapse} use
predictive uncertainty as an OOD signal. These methods are
model-aware, but the resulting scores are usually global in the
feature space and do not explicitly separate departures in predictive
variables from departures in nuisance variables.

\subsection{Tabular, geometric, and calibrated detectors}

A separate line of work screens for unusual test inputs using the
covariate distribution alone. Standard tabular baselines include
distance based outlier scores and $k$-nearest-neighbor criteria
\citep{knorr1998distance,ramaswamy2000outliers}, local outlier factor
\citep{breunig2000lof}, one-class support vector machines
\citep{scholkopf2001estimating}, and tree partition scores such as
Isolation Forest \citep{liu2008isolationforest}. These methods are
useful general anomaly detectors, but they do not use the supervised
response and therefore do not distinguish departures in variables that
matter for prediction from departures in variables that do not.

Recent tabular and clinical studies highlight the practical
importance of this issue. ODROP trains a separate OOD reject option
for disease onset prediction \citep{tosaki2025odrop}, while the
Chameleons benchmark studies OOD behavior in medical tabular data
\citep{azizmalayeri2025chameleons}. Broader medical OOD reviews also
emphasize the need for accurate and interpretable detection in
clinical settings
\citep{zadorozhny2021medood,zamzmi2024radiologyood}. These studies
show the importance of OOD detection for tabular risk prediction, but
they do not provide a model-aware, subspace-aware score for regression
or survival models.

Conformal methods offer a complementary perspective. Given a
nonconformity score, conformal prediction can provide calibrated
$p$-values, intervals, or testing procedures
\citep{angelopoulos2021gentle,bates2023cpood,
liang2024integrativecp,magesh2023principledOOD,novello2024oodcp}.
Such methods can calibrate OOD decisions once a score has been chosen.
Our focus is different. We design the score itself, and leave
conformal calibration of the proposed score for future work.

\subsection{Positioning}

Most existing methods emphasize one of two properties. Some are
model-aware because the score depends on a fitted predictor trained on
labeled data. Others are subspace-aware because the score is computed
after restricting attention to selected variables or learned feature
directions. In the tabular regression and survival settings considered
here, these properties are rarely combined. Classification and deep
representation methods are often model-aware but are not designed for
continuous outcomes or survival analysis. Classical tabular anomaly
detectors work directly with $X$ and are not model-aware.
Uncertainty based regression methods use the fitted model, but usually
score globally in the full feature space.

The proposed method is designed for this gap. It is model-aware because
the score is built from a supervised forest trained on labeled data,
and subspace-aware because the final comparison is restricted to
variables selected as task relevant. The next section gives the
construction.

\section{Methodology}\label{sec3}

We now give the construction of the proposed method. Given labeled
training data $(x_1,y_1),\ldots,(x_n,y_n)$, we first fit a random
forest~\citep{Breiman2001}. For a test input $x$, the forest is then
used to construct two objects needed for scoring. The first is a
signal set $S \subset \{1,\ldots,d\}$ defining the coordinates on
which $x$ is compared with the training data. The second is a local
reference neighborhood $\nn_K(x)$ defining the training cases to which
$x$ is compared. The final score averages a distance from $x$ to this
neighborhood using only the coordinates in $S$.

Random forests are a natural choice for our approach because they
provide a common framework for regression, classification, and
survival outcomes, and because their learned partitions group
observations with similar outcome behavior. This partition based
notion of similarity is commonly summarized by random forest
proximity, where two inputs are considered similar if they often fall
in the same terminal nodes across trees.  Our construction however uses the
forest differently. Rather than using terminal node co-occurrence
itself as the OOD score, we use the learned rule structure of the
forest to build the signal set and the local reference neighborhood.

Each tree can be written as a collection of decision rules. A rule is
a path from the root to a terminal node, expressed as simple
conditions on individual variables, for example $x^{(1)} > 3$ and
$x^{(5)} < 1$. These conditions define a rectangular region of the
input space. Across the ensemble, the full collection of rules
describes how regions of the covariate space are associated with
outcome patterns.  The rules are used in two connected steps. First,
they are analyzed to identify variables carrying predictive
information, yielding a signal set $S \subset \{1,\ldots,d\}$. Second,
for a test input $x$, the rules that contain $x$ are relaxed one
selected variable at a time to identify training points that
repeatedly co-occur with $x$ under the forest. These
co-occurrences rank the training points and form the reference
neighborhood $\nn_K(x)$. The final OOD score is the average distance
from $x$ to this neighborhood, restricted to the signal coordinates,
$d(x) =  \sum_{z \in \nn_K(x)} D_S(x,z)/K$.

The remainder of this section develops each component in turn. A
standalone summary is given in Algorithm~\ref{OutPro}, which we call
\emph{OutPro} (OOD using variable priority). For a test input
$x=(x^{(1)},\ldots,x^{(d)})\in\RR^d$, OutPro takes the labeled
training data and trained forest as input and returns a scalar OOD
score $d(x)$, with larger values indicating greater departure from the
training distribution in the predictive subspace.

\begin{figure}[phtb]
  \centering
    \vspace{0.5in}

%\begin{algorithm*}[!t]
\begin{algorithm}[H]
\caption{OOD using variable priority (OutPro)}
\label{OutPro}
\KwIn{Training data $\{(x_i, y_i)\}_{i=1}^n$; test point $x$; number of neighbors $K$; distance function $D$.}

\KwOut{OOD distance score $d(x)$.}

\vskip10pt

\textbf{Step 1: Train a supervised random forest} \\
Fit a random forest model to $\{(x_i, y_i)\}$ and
extract decision rules $\{\z_j\}$ for terminal nodes. \\

\textbf{Step 2: Apply VarPro to identify signal variables} \\
Obtain the signal set $S = \{s_1, \ldots, s_q\} \subset \{1, \ldots, d\}$ and the importance values $\{w_s\}_{s \in S}$ using VarPro. \\

\textbf{Step 3: Identify rules containing the test point} \\
Let $\rr(x) = \{\z_j : x \in R(\z_j)\}$ be the set of rules whose regions contain $x$. \\

\textbf{Step 4: Compute relative frequencies} \\
\ForEach{training point $x_i$}{
  \ForEach{$s \in S$}{
    Set $n_s(x_i) = 0$. \;
    \ForEach{$\z \in \rr(x)$}{
      Construct the release region $R(\z^s)$ by removing the constraint on variable $s$. \;
      \If{$x_i \in R(\z^s)$}{
        Update $n_s(x_i) \leftarrow n_s(x_i) + 1$.
      }
    }
  }

  Compute $n(x_i) = \sum_{s \in S} n_s(x_i)$. \;
  \If{$n(x_i)=0$}{set $p_s(x_i)=0$ for all $s\in S$ and continue. \;}
  Compute $p_s(x_i) = n_s(x_i) / n(x_i)$ for all $s \in S$. \;
  Compute $G(x_i) = \sum_{s \in S} p_s(x_i) \big(1 - p_s(x_i)\big)$. \;
  Compute $W(x_i) = n(x_i) \cdot G(x_i)$.
}

%\vskip10pt

\textbf{Step 5: Select the $K$ highest-scoring neighbors} \\
Sort $\{x_i\}$ in decreasing order of $W(x_i)$ and select the top $K$
neighbors, $\nn_K(x)$. \\

\textbf{Step 6: Compute the OOD distance} \\
\ForEach{$z \in \nn_K(x)$}{
  Standardize the coordinates using training means and variances. \;
  Compute $D(x, z)$ using the selected distance function and (if
  appropriate)
  the
  normalized importance weights $\{w_s\}_{s \in S}$.
}
Return $d(x) = \frac{1}{K}\sum_{z\in\nn_K(x)} D(x,z)$ for pairwise
metrics, or $d(x) = d_{\text{optics}}(x)$ when OPTICS reachability
is used.

\end{algorithm}

\vspace{1.35in}
\end{figure}

\subsection[Variable Priority and Selection of S]{Variable Priority and Selection of $S$}\label{sec3_varpriority}

Both neighborhood construction and distance evaluation depend on a set
of task relevant variables
$$
S = \{s_1,\ldots,s_q\}\subset\{1,\ldots,d\}.
$$
We estimate $S$, together with weights $\{w_s\}_{s \in S}$ measuring
the relative strength of each selected coordinate, using Variable
Priority (VarPro)~\citep{varpro, ishwaran2025multivariate,
lu2025individual, zhou2026uvarpro}. VarPro is a model based feature
selection method derived from the rule regions produced by a
forest. For a rule region, the release region for variable $s$ is the
expanded region obtained by removing the constraint on $s$ while
keeping all other bounds fixed.

A variable receives high priority when releasing its constraint
consistently changes the outcome behavior of the training points in
the region. Variables whose release has little effect are treated as
noise. Thus VarPro scores variables by outcome changes under rule
relaxation, rather than by split frequency or tree usage alone. This
helps distinguish genuine predictors from correlated surrogates and
is directly relevant for OOD scoring. Including noise variables in
$S$ dilutes the distance calculation across uninformative dimensions,
whereas excluding signal variables can hide departures in the
predictive subspace.

\begin{figure}[phtb]
  \centering

  \vskip-10pt

  \centerline{\emph{Region is released on $x^{(1)}$}}
  \hspace*{-20pt}
  \resizebox{1.95in}{!}{\includegraphics[page=1]{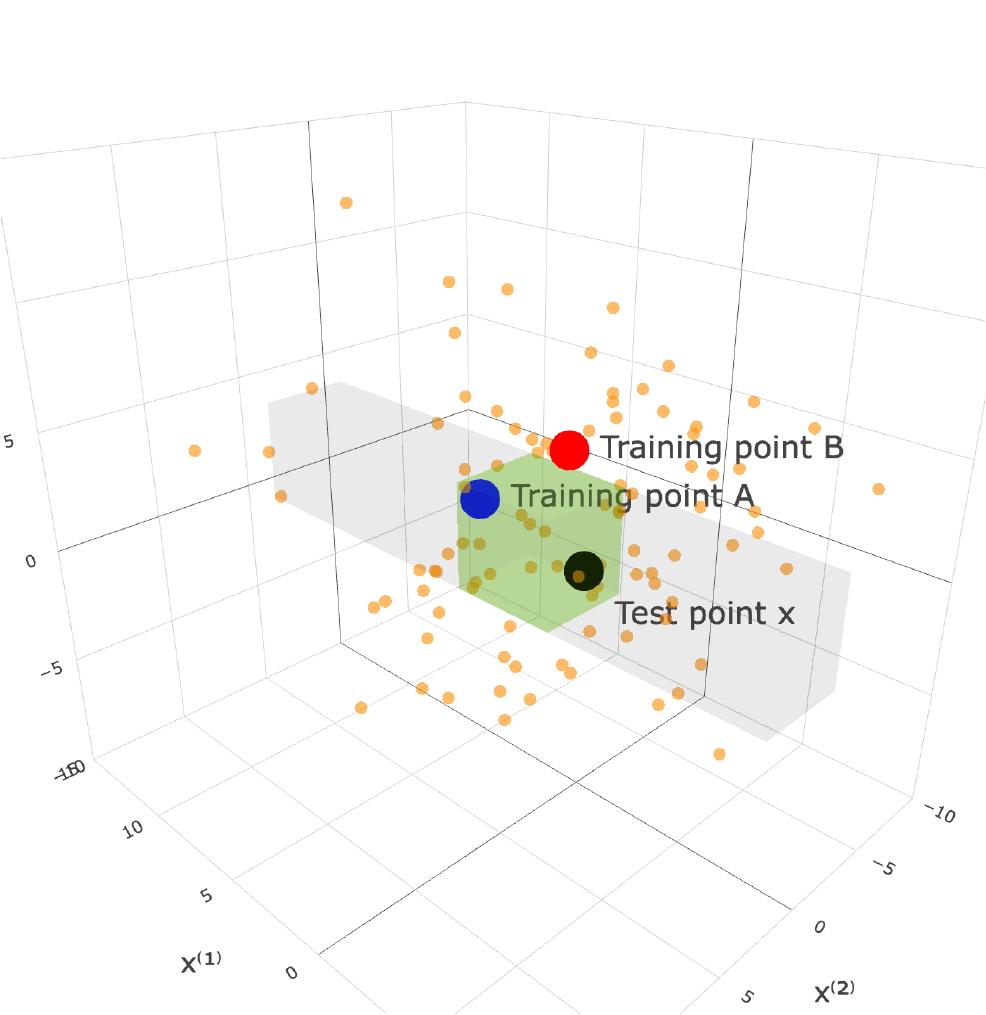}}
  %\hspace*{-20pt}
  \resizebox{2.15in}{!}{\includegraphics[page=1]{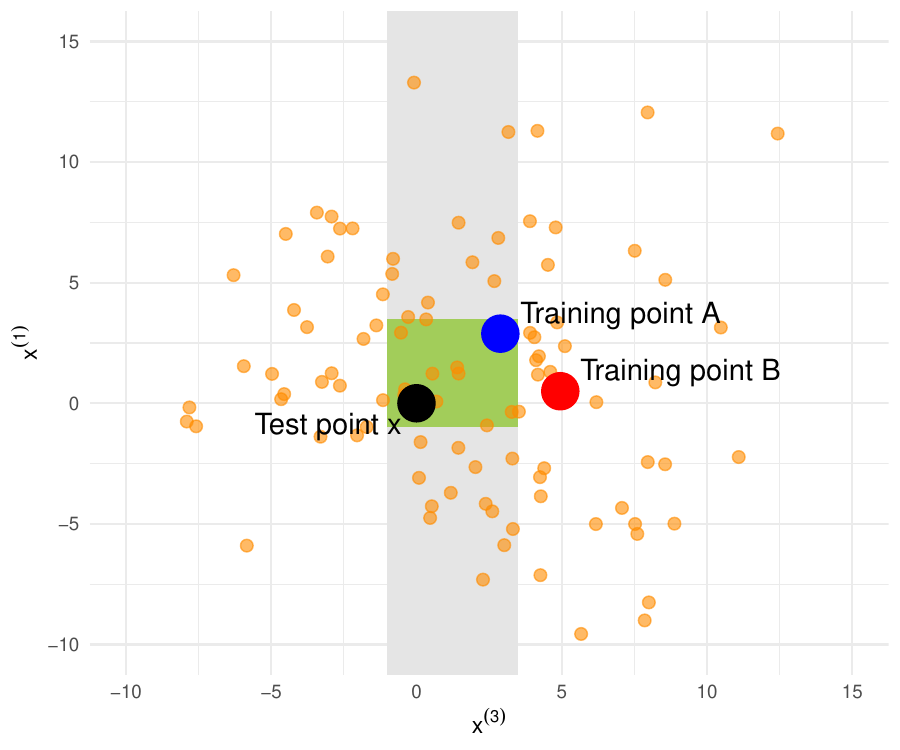}}

  \vskip3pt
  \centerline{\emph{Region is released on $x^{(2)}$}}
  \hspace*{-20pt}
  \resizebox{1.95in}{!}{\includegraphics[page=1]{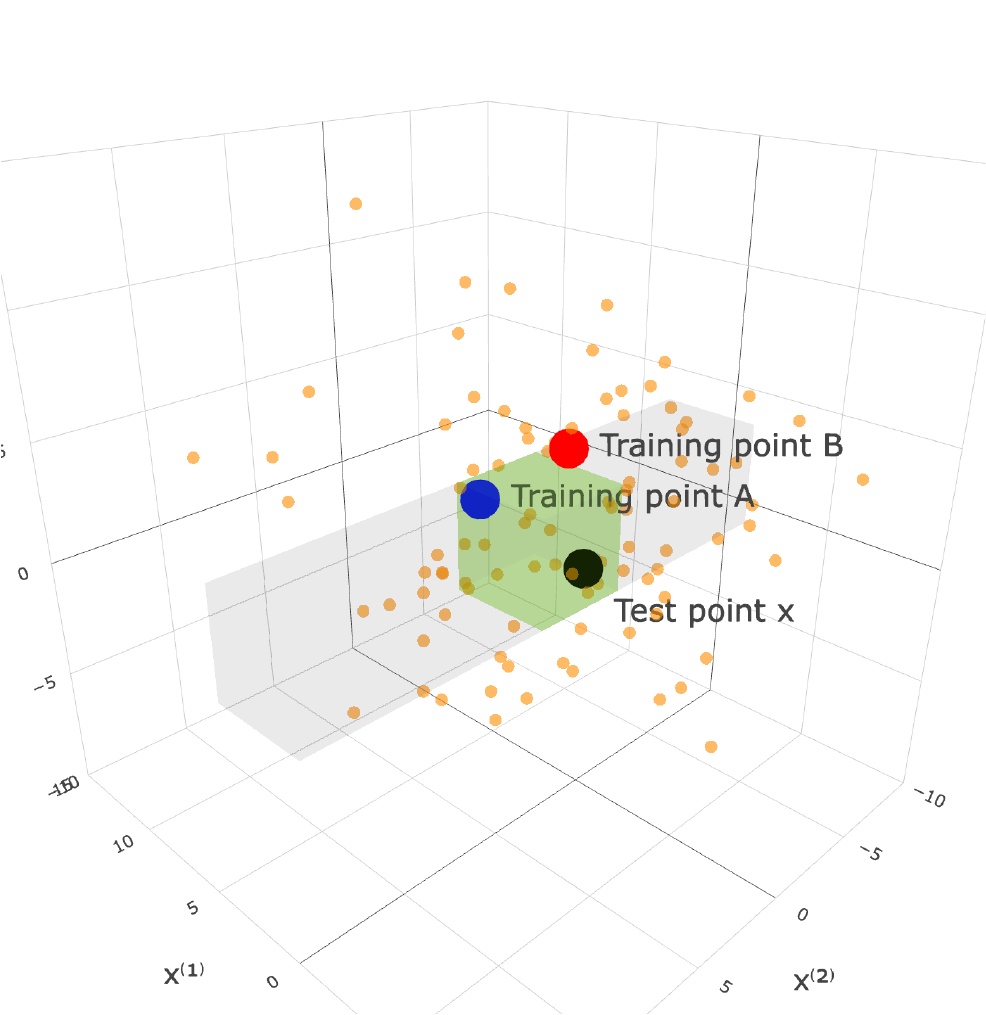}}
  %\hspace*{-20pt}
  \resizebox{2.15in}{!}{\includegraphics[page=2]{neighborillustrate.pdf}}

  \vskip3pt
  \centerline{\emph{Region is released on $x^{(3)}$}}
  \hspace*{-20pt}
  \resizebox{1.95in}{!}{\includegraphics[page=1]{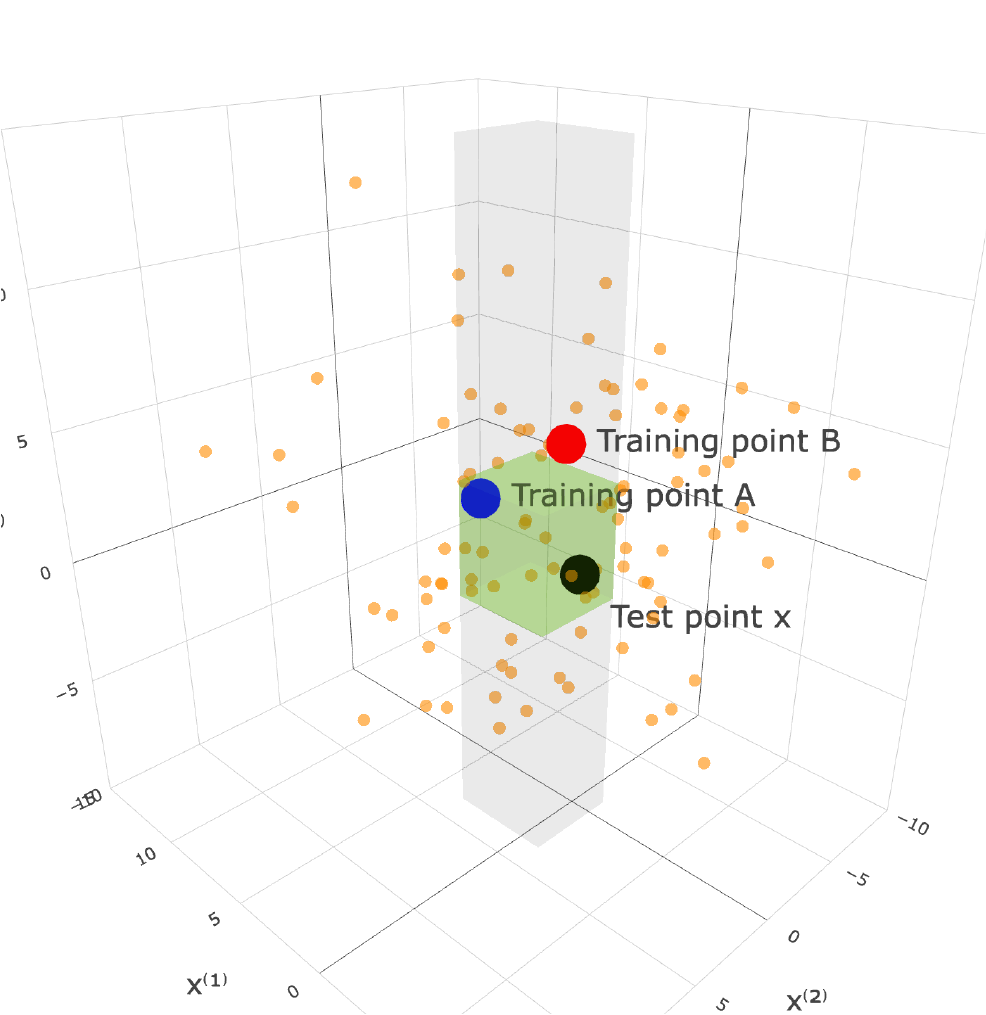}}
  %\hspace*{-20pt}
  \resizebox{2.15in}{!}{\includegraphics[page=3]{neighborillustrate.pdf}}

  \vskip0pt

    \caption{\it Release-region mechanism used for frequency profiling.
    The green cube is a rule containing the test point $x$ shown in
    black, and grey cuboids are release regions formed by removing one
    coordinate constraint at a time. Training points A and B are
    equidistant from $x$ in ordinary Euclidean distance, but their
    release profiles differ. A appears in all three release regions,
    giving counts $(1,1,1)$, whereas B appears only after releasing
    $x^{(2)}$ and $x^{(3)}$, giving counts $(0,1,1)$. Thus A is more
    similar to $x$ under the rule relaxation profile.}

  \label{neighborillustrate}
\end{figure}

\subsection{Frequency Profiling via Rule Relaxation}\label{sec3_freq}

The same release operation is used to build the reference neighborhood
for a test input $x$. Let $\rr(x)$ denote the set of decision rules
from the forest whose terminal regions contain $x$. Each rule
$\z \in \rr(x)$ defines a hyperrectangle
$R(\z) \subset \RR^d$. For each selected coordinate $s \in S$, let
$R(\z^s)$ be the release region formed by removing the constraint on
$s$ and retaining all other bounds.

For each training point $x_i$, we count its appearances in these
release regions. For $s \in S$, define
$$
n_s(x_i) = \sum_{\z \in \rr(x)} I\{x_i \in R(\z^s)\},
$$
and let
$$
n(x_i) = \sum_{s \in S} n_s(x_i).
$$
When $n(x_i)>0$, the frequency profile of $x_i$ relative to $x$ is
$$
(p_{s_1}(x_i), \ldots, p_{s_q}(x_i)) =
\left( \frac{n_{s_1}(x_i)}{n(x_i)}, \ldots,
       \frac{n_{s_q}(x_i)}{n(x_i)} \right).
$$
If $n(x_i)=0$, we set $p_s(x_i)=0$ for all $s \in S$ and exclude
$x_i$ from neighborhood construction. The profile records how often a
training point co-occurs with $x$ after releasing each selected
coordinate, and therefore summarizes its alignment with $x$ under the
forest's local rule structure.

\autoref{neighborillustrate} illustrates this construction for a
single rule and three selected coordinates. Points A and B are
equidistant from $x$ in ordinary Euclidean distance, but their release
profiles differ. Point A appears in all three release regions, whereas
point B appears in only two. The frequency profile therefore captures
local similarity under the forest's rule structure, rather than under
a fixed Euclidean metric.

\subsection{Proximity Scoring from Frequency Dispersion}\label{sec3_prox}

This subsection defines a proximity score $W(x_i)$ for each training
point $x_i$, relative to the test input $x$. The score is used to rank
training points before the final OOD distance is computed.  A training
point receives a large value of $W(x_i)$ when it appears often in the
release regions for $x$ and when those appearances are spread across
the selected variables in $S$. A point receives a smaller value when
it appears rarely, or when its appearances are concentrated in release
regions for only a few selected variables.

For a training point $x_i$ with $n(x_i)>0$, recall that
$p_s(x_i)$ is the relative frequency with which $x_i$ appears after
releasing coordinate $s \in S$. We summarize the balance of this
profile using the Gini impurity
$$
G(x_i) = \sum_{s \in S} p_s(x_i) \big(1 - p_s(x_i)\big).
$$
Larger values occur when co-occurrence is spread more evenly across
the selected coordinates. The final proximity score also accounts for
the total number of appearances,
$$
W(x_i) = n(x_i) G(x_i).
$$
Thus $W(x_i)$ favors training points that both appear frequently in
release regions and have balanced co-occurrence across the predictive
coordinates.

\subsection{OOD Distance via Subspace Distance Functions}\label{sec3_dist}

The reference neighborhood for $x$ is the set $\nn_K(x)$ of the
$K$ training points with the largest proximity scores $W(x_i)$.
The OOD score is then the average distance from $x$ to this
neighborhood,
$$
d(x) = \frac{1}{K} \sum_{z \in \nn_K(x)} D(x,z).
$$
The distance $D(x,z)$ is computed only on the selected variables
$s \in S$ and is weighted by the VarPro scores $\{w_s\}_{s \in S}$.
The weights are positive and normalized, so higher priority variables
receive greater emphasis and lower priority variables are
down-weighted.  Several different types of distance functions are used
in the empirical experiments, these are conveniently 
summarized in
Table~\ref{tab:subspace_distances}.

\begin{figure}
\centering  
\vspace*{1in}

%\begin{table}
\small
\renewcommand{\arraystretch}{1.35}
\captionof{table}{\it Distance functions used in the experiments. Coordinates
  are standardized using training means and variances, and all
  distances are restricted to the selected coordinates $S$. Except for
  Mahalanobis distance, each coordinate $s\in S$ is weighted by its
  normalized VarPro weight $w_s$. The first five rows define
  pairwise distances $D(x,z)$ used in the averaged OOD score, while
  OPTICS defines a direct reachability score for $x$.}
\label{tab:subspace_distances}
\begin{tabular}{p{0.22\linewidth}p{0.70\linewidth}}
\hline
Distance & Definition \\
\hline

Product
&
$\displaystyle
D_{\text{prod}}(x,z)
=
\prod_{s \in S}
\left(|x^{(s)}-z^{(s)}|+\varepsilon\right)^{w_s}
$,
with $\varepsilon>0$ used for numerical stability.
\\

Euclidean
&
$\displaystyle
D_{\text{euclid}}(x,z)
=
\left\{
\sum_{s \in S} w_s (x^{(s)}-z^{(s)})^2
\right\}^{1/2}.
$
\\

Manhattan
&
$\displaystyle
D_{\text{manh}}(x,z)
=
\sum_{s \in S} w_s |x^{(s)}-z^{(s)}|.
$
\\

Minkowski
&
$\displaystyle
D_{\text{mink}}(x,z)
=
\left\{
\sum_{s \in S} w_s |x^{(s)}-z^{(s)}|^p
\right\}^{1/p}.
$
We use $p=4$ in the empirical studies.
\\

Mahalanobis
&
Let $\Sigma^{(S)}$ be the empirical covariance matrix of the training
data restricted to $S$. Then
$\displaystyle
D_{\text{mahal}}(x,z)
=
\left\{
(x^{(S)}-z^{(S)})^T
(\Sigma^{(S)})^{-1}
(x^{(S)}-z^{(S)})
\right\}^{1/2}.
$
\\

OPTICS
&
OPTICS~\citep{ankerst1999optics} is applied to
$\nn_K(x)\cup\{x\}$ after restricting to $S$ and scaling coordinate
$s$ by $\sqrt{w_s}$. The OOD score is the reachability value
$\displaystyle
d_{\text{optics}}(x)
=
\text{Reachability}
\big(x \mid \nn_K(x), \texttt{minPts}, \{w_s\}_{s \in S}\big).
$
We use the \texttt{optics} function from the R package
\texttt{dbscan} with \texttt{minPts}=5
\citep{hahsler2019dbscan}.
\\

\hline
\end{tabular}
%\end{table}

\vspace*{1in}
\end{figure}

\section{Comparison Methods}\label{sec4}

We compare OutPro with OOD detectors that are applicable to
continuous outcomes and have reliable open source implementations.
The baselines include uncertainty based methods, latent representation
methods, gradient based scores, and classical unsupervised tabular
detectors. The unsupervised methods use only the training covariates,
whereas the model based methods use a fitted predictor.
Table~\ref{tab:comparison_methods} summarizes the score and
implementation used by each comparison method.

All neural network methods share the same base architecture,
implemented with the Keras API~\citep{chollet2015keras}. The network
has two hidden ReLU layers, batch normalization, dropout, and a single
output node trained by mean squared error. MSP, ODIN, Energy,
Sensitivity, DKL, GMM, and latent space $k$-nearest neighbors are
computed from this shared fitted network, as described in
Table~\ref{tab:comparison_methods}. Conformal prediction and deep
ensembles use the same architecture but refit it in their own split
calibration and ensemble settings. Gaussian process methods use the
\texttt{laGP} package~\citep{pkg:laGP,laGP}. All scores are oriented
so that larger values indicate stronger evidence of OOD behavior.

\begin{table}[phtb]
\centering
\small
\renewcommand{\arraystretch}{1.20}
\caption{Comparison methods. Here $f(x)$ is the fitted prediction,
$h(x)$ is the penultimate layer representation of the neural network,
$\bar y_{\mathrm{train}}$ is the mean training response, and
$\tilde x=x+\epsilon\operatorname{sign}(x)$ with $\epsilon=0.01$.
Input space methods use standardized covariates unless noted
otherwise.}
\label{tab:comparison_methods}
\begin{tabular}{p{0.25\linewidth}p{0.67\linewidth}}
\hline
Method & OOD score or implementation summary \\
\hline
MSP~\citep{hendrycks2016baseline}
&
Regression adaptation using prediction deviation,
$|f(x)-\bar y_{\mathrm{train}}|$. This is monotone equivalent to the
reciprocal confidence score used in the implementation.
\\

ODIN~\citep{liang2018odin}
&
Regression adaptation using the perturbed input $\tilde x$ and score
$|f(\tilde x)-\bar y_{\mathrm{train}}|$.
\\

Energy~\citep{liu2020energy}
&
Perturbation based latent score $\|h(\tilde x)\|_2^2$.
\\

Mahalanobis
&
Input space Mahalanobis distance
$\{(x-\mu)^T\Sigma^{-1}(x-\mu)\}^{1/2}$, where $\mu$ and $\Sigma$ are
estimated from the training covariates.
\\

LOF~\citep{breunig2000lof}
&
Local outlier factor computed on standardized covariates with
$k=\min(20,n-1)$.
\\

Isolation Forest~\citep{liu2008isolationforest}
&
Isolation Forest anomaly score using 200 trees and subsample size
$\min(256,n)$.
\\

OCSVM~\citep{scholkopf2001estimating}
&
One class support vector machine with radial kernel
$K(x,x')=\exp\{-\gamma\|x-x'\|_2^2\}$, $\gamma=1/d$. The OOD score is
the negative fitted decision function. The parameter $\nu$ is set to
the benchmark cutoff rate, truncated to
$[1/\max(2,n),0.5]$.
\\

kNN density~\citep{knorr1998distance,ramaswamy2000outliers}
&
Robust $k$-nearest-neighbor density score. Covariates are centered by
the training median and scaled by the training median absolute
deviation. The score is the log median radius to the
$k=\min(20,n-1)$ nearest training points.
\\

Conformal prediction~\citep{angelopoulos2021gentle}
&
Split conformal score using 75\% of the training data for fitting and
25\% for calibration. The ranking score is
$|f(x)-\operatorname{median}(f_{\mathrm{calib}})|$; the binary
decision compares this value with the corresponding calibration
quantile.
\\

Gaussian process
&
Local approximate Gaussian process regression using \texttt{laGP};
the OOD score is the posterior predictive variance at $x$.
\\

Sensitivity~\citep{he2018softmax}
&
Average absolute input gradient,
$d^{-1}\sum_{j=1}^d |\partial f(x)/\partial x^{(j)}|$.
\\

DKL~\citep{wilson2016dkl}
&
Deep kernel learning baseline. A Gaussian process is fit by
\texttt{laGP} to the neural representation $h(x)$, and the OOD score
is the posterior predictive variance.
\\

Deep ensembles~\citep{Lakshminarayanan2017DeepEnsembles}
&
Variance of predictions across independently initialized neural
network ensemble members.
\\

GMM~\citep{Pleiss2019OODRegression}
&
Gaussian mixture density in PCA reduced latent space. The OOD score
is $-\log p(\tilde h(x))$, where $\tilde h(x)$ is the reduced
representation.
\\

DkNN~\citep{sun2022deepknn}
&
Latent space $k$-nearest-neighbor score,
$k^{-1}\sum_{j=1}^k\|h(x)-h(x_{(j)})\|_2$, using nearest neighbors in
the neural representation.
\\
\hline
\end{tabular}
\end{table}

\section{Benchmarking OOD Detection Methods in Regression}\label{sec5}

We evaluate performance using a combination of simulated and real
regression problems.  On synthetic data, we use a well-known machine
learning model with independent covariates
(Section~\ref{sec5.fr}), which provides a controlled anomaly framework
with clearly defined ground-truth labels.  For real data, we consider
two sets of experiments.  The first is a large collection of
regression benchmarks (Section~\ref{sec5.pblmr}) with varying sample
sizes, dimensions, and signal-to-noise ratios.  The second uses
high-dimensional microarray survival studies
(Section~\ref{sec5.micro}), where the number of features greatly
exceeds the sample size.  For the real-data experiments, anomalies are
generated using a copula-based strategy described in
Section~\ref{sec5.copula}, which allows targeted perturbations of
marginal distributions and joint dependence.  Test cases are drawn
from a held-out subset of the data, and anomalous inputs are created
by perturbing selected features to simulate OOD behavior.

\subsection{Friedman Simulation}\label{sec5.fr}

We first consider a simulated regression setting based on the
well-known Friedman model~\citep{Friedman1991}.  Each covariate
$X^{(j)} \sim U[0,1]$ independently for $j = 1, \ldots, d$.  The
response is generated according to
$$
Y = 10 \sin\big(\pi X^{(1)} X^{(2)}\big) + 20 \big(X^{(3)} -
0.5\big)^2 + 10 X^{(4)} + 5 X^{(5)} + \varepsilon,
$$
where $\varepsilon \sim N(0, \sigma^2)$ and the remaining $d - 5$
covariates are irrelevant noise features.  We
use $d = 20$ and $\sigma = 1$.

To simulate OOD instances, we apply a fixed additive shift to a subset
of features identified as predictive of the outcome.  Although the
true signal variables are known in this simulation, we use tree-based
gradient boosting~\citep{Friedman2001} to rank features by their
contribution to predictive accuracy.  This data driven approach to
feature selection mirrors the methodology used in our later real-data
experiments and ensures consistency across settings.  From the ranked
list, the top ten percent of features are selected, and each is
perturbed by a fixed additive amount scaled by its standard deviation,
with direction randomly assigned for each test point.

\subsubsection{Ground Truth}

The additive shift procedure is a common strategy in OOD detection
research.  However, the labeling rule that defines a shifted point as
anomalous has a subtle issue that is often overlooked: a shifted point
may lie entirely within the support of the training distribution and
therefore represent a perfectly valid input.

To see this, note that under the Friedman setup the covariates
$X^{(j)}$ are independent and uniformly distributed on $[0,1]$.
Applying an additive shift
$\delta = (\delta_1, \ldots, \delta_d)$ to the covariate vector
$X = (X^{(1)}, \ldots, X^{(d)})$ produces a perturbed point
$$
X^* = X + \delta.
$$
Because the covariates are independent with bounded support, a shifted
point $X^*$ becomes anomalous only if at least one coordinate falls
outside the unit interval.  For coordinate $j$, the condition
$X^{(j)} + \delta_j \in [0,1]$ is equivalent to
$$
-\delta_j \le X^{(j)} \le 1 - \delta_j.
$$
Hence the probability that all coordinates remain within the original
hypercube is
$$
\PP(\text{all coordinates inside}) = \prod_{j=1}^d (1 - |\delta_j|)_+, 
$$
where $(a)_+ := \max(a, 0)$.  The true fraction of shifted points
that fall outside the support is therefore $1$ minus this value, which
for small shifts can be approximated to first order as
$\sum_{j=1}^d |\delta_j|$ and can therefore be substantially less
than~$1$.

Therefore many nominally shifted points may still lie entirely
within the original support and are therefore legitimate data values,
particularly when shifts are small or the dimension is moderate.  To
avoid this ambiguity, we adopt the rule that a point is labeled
anomalous if and only if at least one of its coordinates falls outside
$[0,1]$.

\subsubsection{Experimental Settings}

We generated $100$ independent datasets of size $n = 2000$ from the
Friedman model.  Each dataset was randomly split into 80\% training
and 20\% testing and anomalies were created by perturbing test points
using shifts as described above.  Shifts were applied to the top
predictive features ranked by gradient boosting and scaled by the
empirical standard deviation of each feature.  We considered five
shift magnitudes: $0.05$, $0.25$, $0.5$, $1.0$, and $2.0$.
Ground-truth labels were defined by classifying a perturbed test point
as anomalous if and only if at least one coordinate fell outside
$[0,1]$.

\subsubsection{Results}

\begin{figure}[phtb!]
  \centering
  \resizebox{5.0in}{!}{\includegraphics[page=1]{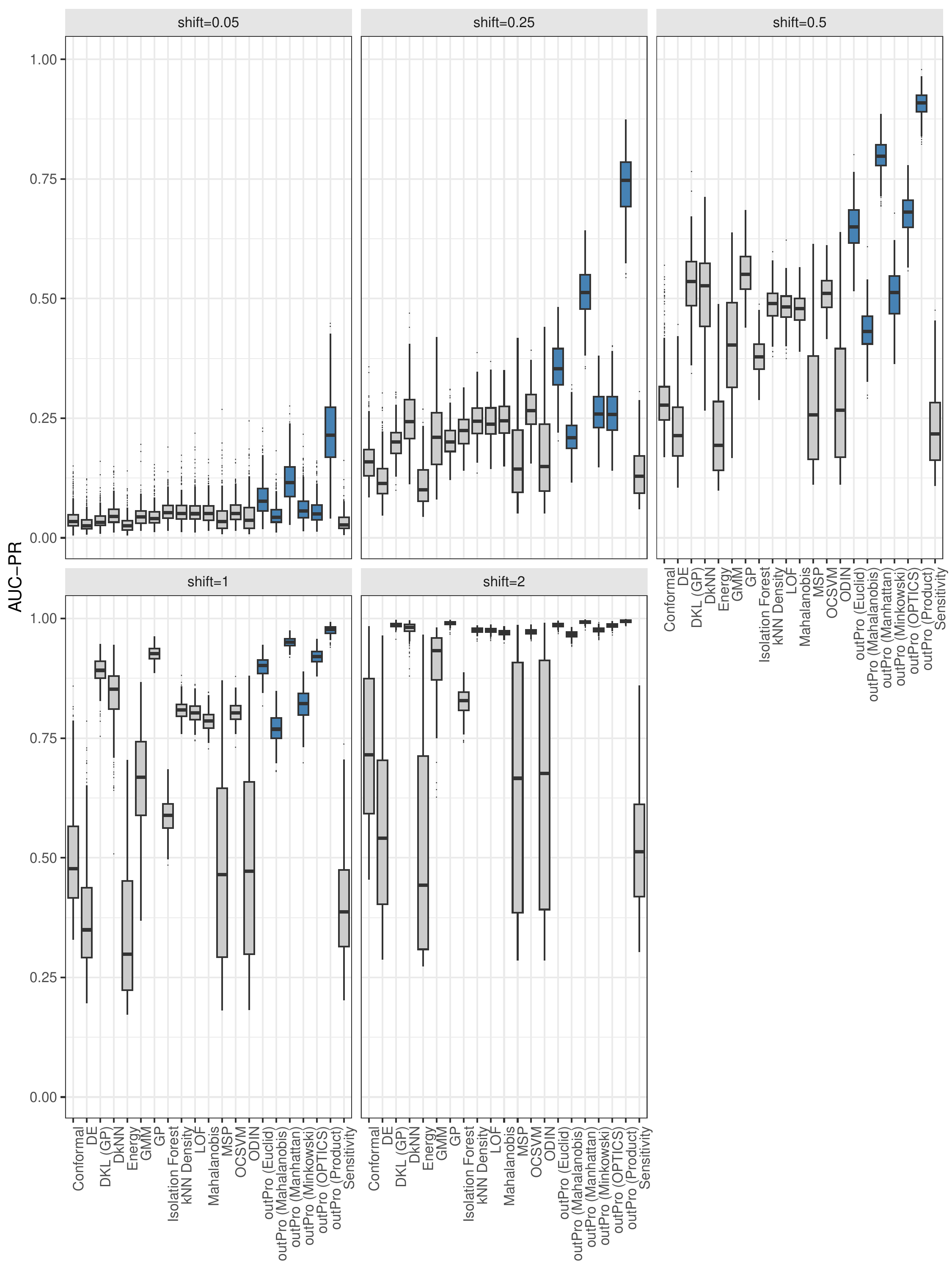}}

  \vskip15pt

  \caption{\it Boxplots of AUC-PR values for each method across 100
    runs at varying shift magnitudes in the Friedman simulation. Blue
    indicates OutPro methods. Performance improves for all methods as the
    shift magnitude increases. The OutPro product and Manhattan scores
    are best for small shifts ($0.05$ to $0.5$), where detection is
    most difficult.}

  \label{friedman.box}
\end{figure}

\begin{figure}[phtb!]
  \centering
  \resizebox{5.0in}{!}{\includegraphics[page=1]{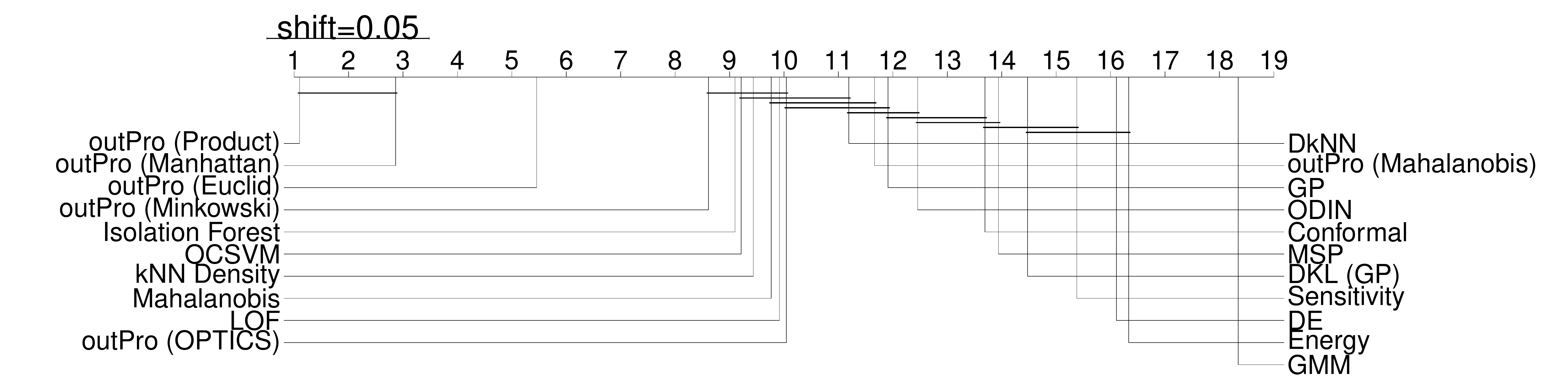}}\\[8pt]
  \resizebox{5.0in}{!}{\includegraphics[page=2]{benchmark_friedman_cd.pdf}}\\[8pt]
  \resizebox{5.0in}{!}{\includegraphics[page=3]{benchmark_friedman_cd.pdf}}\\[8pt]
  \resizebox{5.0in}{!}{\includegraphics[page=4]{benchmark_friedman_cd.pdf}}\\[8pt]
  \resizebox{5.0in}{!}{\includegraphics[page=5]{benchmark_friedman_cd.pdf}}\\[8pt]

  %\vskip10pt

  \caption{\it Critical difference (CD) plots comparing average AUC-PR
    rankings for the Friedman simulation at each shift magnitude. Lower
    ranks indicate better performance. Methods with no statistically
    significant difference at $\alpha = 0.05$ are connected by a
    horizontal bar. For shifts $0.05$, $0.25$, and $0.5$, the OutPro
    product and Manhattan scores rank near the top, while the
    best classical tabular baselines are in the next tier. The
    product score has the best average rank at every shift
    magnitude.}

  \label{friedman.cd}
\end{figure}

Performance was measured by the area under the precision-recall curve
(AUC-PR). Boxplots in \autoref{friedman.box} show that the OutPro
procedures are among the best for small shifts ($0.05$ to $0.5$),
which is the setting where anomalies are subtle and remain close to
the training distribution.  To systematically compare methods,
critical difference (CD) plots are given in \autoref{friedman.cd},
with significance assessed using the Nemenyi test. The OutPro product
score attains the best average rank at every shift magnitude, and
Manhattan is consistently second. For the smallest shifts, the best
baselines are the classical tabular methods, especially Isolation
Forest, OCSVM, robust $k$NN density, LOF, and the global Mahalanobis
score, but they do not overtake OutPro. As the shift grows, GP and DKL
(GP) improve and, together with DkNN, become competitive with OutPro.
Isolation Forest is less effective for the larger shifts, while robust
$k$NN density and LOF remain in the middle. Overall, the results show
that density based tabular baselines can be competitive when the shift
is large enough to create obvious low density points, but OutPro is
best when the perturbation is small and functionally targeted.

\subsection{Copula-Based Strategy for Anomaly Generation}\label{sec5.copula}

The Friedman simulation provides a controlled benchmark with
independent features and bounded support. Real tabular data are more
complex, with skewed or heavy-tailed marginals and nontrivial
dependence among variables. To generate realistic anomalies in this
setting, we use a copula-based strategy that separates marginal
behavior from dependence and perturbs each component in a controlled
way.

Let $X=(X^{(1)},\ldots,X^{(d)})$ be a continuous random vector with
marginal CDFs $F_1,\ldots,F_d$. Applying the probability integral
transform coordinatewise gives
$$
U^{(j)} = F_j(X^{(j)}), \quad j=1,\ldots,d.
$$
For continuous marginals, each $U^{(j)}$ is uniform on $[0,1]$, and
the joint distribution of $U=(U^{(1)},\ldots,U^{(d)})$ is the copula
$C$. By Sklar's theorem~\citep{sklar1973random},
$$
F(x^{(1)},\ldots,x^{(d)})
=
C\{F_1(x^{(1)}),\ldots,F_d(x^{(d)})\}.
$$
Thus a point on the copula scale can be mapped back to the data scale
by $x^{(j)}=F_j^{-1}(u^{(j)})$. The marginal inverse CDFs determine
the coordinate scales, while the joint arrangement of the $u$ values
determines dependence.

\begin{figure}[phtb!]
  \centering
  \resizebox{5in}{!}{\includegraphics{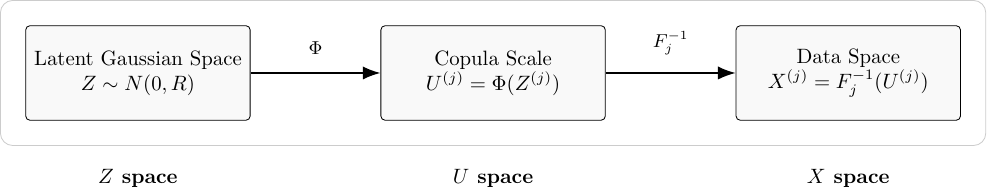}}
  
  \caption{\it Three-stage transformation used by the copula anomaly
  generation procedure. A latent Gaussian vector $Z$ determines
  dependence, the transformation $U = \Phi(Z)$ places each coordinate
  on a common uniform scale, and the inverse marginal CDFs map back to
  the data space $X$. Perturbing different stages of this pipeline
  yields the \warp, \joint, and \support\ anomaly modes.}

  \label{copula}
\end{figure}

We model dependence with a Gaussian copula
\citep{joe2014dependence,nelsen2006introduction}. A latent Gaussian
vector $Z=(Z^{(1)},\ldots,Z^{(d)})$ is generated as
$Z \sim N(0,R)$, where $R$ is estimated from the training data, and
$U^{(j)}=\Phi(Z^{(j)})$, $j=1,\ldots,d$.
This gives the three-stage transformation (\autoref{copula}) 
$$
Z \longrightarrow U \longrightarrow X,
$$
where $Z$ controls dependence, $U$ places coordinates on a common
uniform scale, and the inverse marginals map back to the observed data
space. Perturbing different stages of this pipeline yields the three
anomaly modes summarized in
Table~\ref{tab:copula_modes}.

The three anomaly modes are designed
to test different ways a tabular distribution can depart from the
training data. The \warp\ mode changes marginal tail behavior by
distorting the uniform variables before mapping back to the original
coordinate scales. This produces observations that remain tied to the
estimated dependence structure but have altered marginal behavior. The
\joint\ mode perturbs the latent Gaussian representation, creating
points in atypical regions of the dependence structure while
preserving the original marginal mapping. The \support\ mode moves a
latent training point outward and then maps it through modified
inverse CDFs that extrapolate beyond the observed coordinate ranges.

\begin{table}[phtb]
\centering
\small
\renewcommand{\arraystretch}{1.25}
\caption{\it Copula anomaly modes used in the experiments. Here
$z_i=\Phi^{-1}\{F(x_i)\}$ denotes the latent Gaussian representation
of training point $x_i$, $q$ is the tail probability, and
$\tau_{1-q}$ is the empirical $(1-q)$ percentile of
$d_M(x_i)=\{z_i^T R^{-1} z_i\}^{1/2}$ over the training data (all
experiments use $q=0.05$).  The value $a^*$ is a binary label
with value 1 for anomalous data.}
\label{tab:copula_modes}
\begin{tabular}{p{0.13\linewidth}p{0.78\linewidth}}
\hline
Mode & Generation and labeling \\
\hline

\warp
&
Draw $z\sim N(0,R)$ and set $u^{(j)}=\Phi(z^{(j)})$. Apply the
coordinatewise warp
$
u^{*(j)} =
\{u^{(j)}\}^{\gamma}/
[\{u^{(j)}\}^{\gamma}+\{1-u^{(j)}\}^{\gamma}]
$
with $\gamma>1$, and map back by
$x^{*(j)}=F_j^{-1}(u^{*(j)})$. The label is
$a^*=1\{d_M(x^*)>\tau_{1-q}\}$, where
$z^*=\Phi^{-1}(u^*)$ and
$d_M(x^*)=\{z^{*\top}R^{-1}z^*\}^{1/2}$.
\\

\joint
&
Choose a latent training point $z_b$, draw a random unit vector
$\zeta$, and sample
$r=\sqrt Q$ with
$Q\sim\chi^2_d$ truncated to
$[\chi^2_d(1-q),\infty)$. Set
$z^*=z_b+r\zeta$ and map back by
$x^{*(j)}=F_j^{-1}\{\Phi(z^{*(j)})\}$. These points perturb the
dependence structure while preserving the marginal mapping, and are
labeled anomalous, $a^*=1$.
\\

\support
&
Choose a latent training point $z_b$ with
$r_b=\{z_b^T R^{-1}z_b\}^{1/2}$. Draw
$r^*=\sqrt{Q^*}$ from the upper $(1-q)$ tail of $\chi^2_d$ and set
$z^*=z_b(r^*/r_b)$. Map back using a modified inverse CDF,
$x^{*(j)}=\tilde F_j^{-1}\{\Phi(z^{*(j)})\}$, which extrapolates
beyond the empirical support. These points are labeled anomalous, $a^*=1$.
\\

\hline
\end{tabular}
\end{table}

\subsection{Benchmarking on Diverse Real and Simulated Datasets}\label{sec5.pblmr}

For benchmarking we used a curated subset of regression datasets from
the Penn Machine Learning Benchmark (PMLB)
repository~\citep{olson2017pmlb, pkg:pmlbr}.  Only datasets with at
least 10 features and a continuous response were retained, resulting
in 61 datasets with dimensions ranging from $d=10$ to $124$ and sample
sizes from $n=47$ to $1066$.

\subsubsection{Experimental Settings}

Each dataset was randomly split into 80\% training and 20\% testing.
Synthetic anomalies were generated using the copula-based procedure
described in Section~\ref{sec5.copula}, with the number of
anomalies matched to the number of test points. Each copula mode
(\warp, \joint, \support) was applied independently. To ensure that
perturbations produced meaningful distributional shifts, the copula
transformations were applied only to features identified as
predictive of the outcome by selecting the top 10\% using gradient
boosting, as in the Friedman simulation, while the remaining features
were left unchanged. Performance was measured by AUC-PR, with all
evaluations repeated over 100 independent runs.

\subsubsection{Results}

\begin{figure}[phtb!]
  \centering
  %\vskip1in
  \resizebox{5.25in}{!}{\includegraphics[page=1]{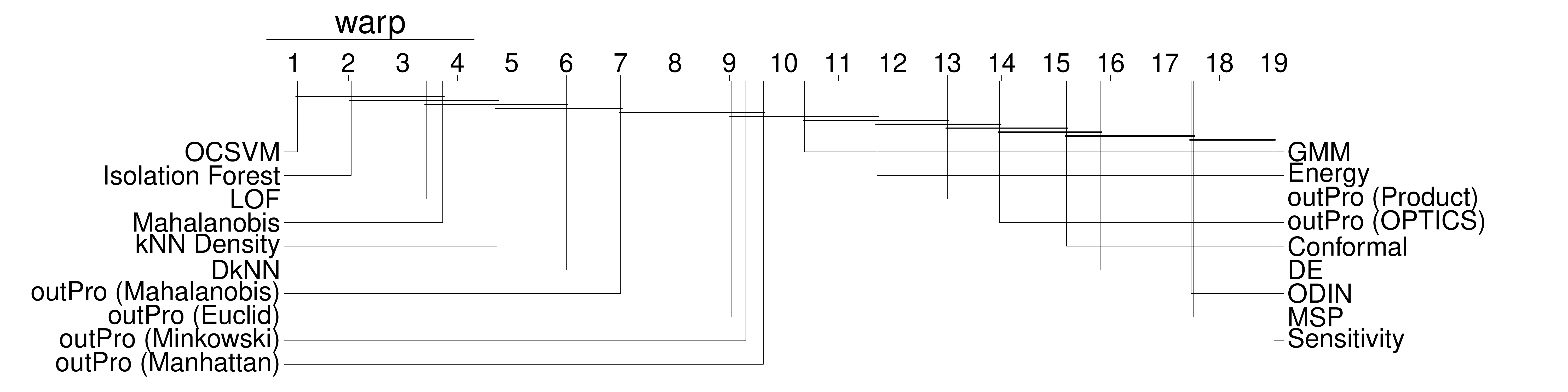}}\\[10pt]
  \resizebox{5.25in}{!}{\includegraphics[page=2]{benchmark_pmlbr.pdf}}\\[10pt]
  \resizebox{5.25in}{!}{\includegraphics[page=3]{benchmark_pmlbr.pdf}}\\[10pt]

  %\vskip10pt

  \caption{Critical difference (CD) plots comparing OOD detection
    methods across 61 PMLB regression datasets. From top to bottom, the
    panels correspond to the three anomaly generation modes, \warp,
    \joint, and \support. Methods are ranked by average AUC-PR, and
    statistically indistinguishable groups at $\alpha = 0.05$ are
    connected by horizontal bars based on the Nemenyi test.}

  \label{pmlbr}
  %\vskip1in
\end{figure}

Given the large number of datasets, we summarize performance by the
average AUC-PR rank across datasets, shown as critical difference (CD)
plots in \autoref{pmlbr}. Under the \warp\ mode, the best methods are
OCSVM, Isolation Forest, LOF, the global Mahalanobis score, robust
$k$NN density, and DkNN.  Next are the OutPro procedures, with
Mahalanobis performing the best, followed by Euclidean, Minkowski, and
Manhattan, while the product and OPTICS versions are lower. This
pattern is reasonable because \warp\ creates marginal tail distortion
within the observed support and therefore favors methods driven by
density, isolation, or global geometry.

Under the \joint\ mode, the pattern changes significantly. The OutPro
procedures are the best, with Manhattan first, followed by Euclidean,
product, Minkowski, OPTICS, and Mahalanobis. The best non-OutPro
methods are OCSVM, LOF, with energy, robust $k$NN density, and the
global Mahalanobis score forming the next group. This shows that when
anomalies alter dependence, methods built from local predictive
profiles perform better than generic full-space scoring rules.

The same pattern holds for the \support\ mode. OutPro procedures
are again the best.  Among the non-OutPro methods, global Mahalanobis
score is the best, followed by OCSVM, LOF, robust $k$NN density, and
energy, while conformal and Isolation Forest are in the next
group. This shows that when anomalies exist outside the observed
support, methods derived from local predictive profiles are especially
effective.

\subsection{High-Dimensional Benchmarking}\label{sec5.micro}

We next evaluate performance on five high-dimensional microarray datasets
commonly used in the survival analysis literature, diffuse large
B-cell lymphoma (DLBCL)~\citep{Rosenwald2002}, breast
cancer~\citep{vantVeer2002}, lung cancer~\citep{Beer2002}, acute
myeloid leukemia (AML)~\citep{Bullinger2004}, and mantle cell
lymphoma (MCL)~\citep{Rosenwald2003}.

\begin{table}[phtb!]
  \centering
  \caption{\it Summary of high-dimensional microarray datasets used for
  benchmarking after Cox-score filtering.}
  \label{tab:microarray}
  %\vskip10pt
  %\hspace*{1.5in}
  \begin{tabular}{lcc}
    \toprule
    Dataset & $n$ & $d$ \\
    \midrule
    AML              & 116 & 629 \\
    Breast Cancer    &  78 & 475 \\
    DLBCL            & 240 & 740 \\
    Lung Cancer      &  86 & 713 \\
    MCL              &  92 & 881 \\
    \bottomrule
  \end{tabular}
\end{table}

To place these in a regression framework, we converted survival
outcomes into continuous pseudo-responses using random survival forests
(RSF)~\citep{Ishwaran2008}.  RSF models were fit to each dataset, and
out-of-bag (OOB) mortality predictions were extracted for use as
regression targets.  Although OutPro can handle survival data
directly, this transformation enabled direct comparison with the
baseline methods.  To improve the signal-to-noise
ratio, features were filtered by Cox-score
ranking~\citep{bair2004}.  Table~\ref{tab:microarray} summarizes
sample sizes and dimensions for the filtered datasets.

\subsubsection{Experimental Settings}

The experimental design mirrored that of the PMLB analysis. Each
dataset was split into 80\% training and 20\% test sets. Synthetic
anomalies were generated using the copula-based procedure under all
three modes, with the number of anomalies matched to the number of
test points. Anomaly perturbations were restricted to the top 10\%
of features identified as predictive by gradient boosting, while the
remaining features were left unchanged. Each configuration was
repeated over 100 independent runs, and performance was measured
using AUC-PR. Due to the high-dimensional setting, the global
Mahalanobis baseline could not be evaluated directly because the
sample covariance matrices were singular, and therefore is
omitted.

\subsubsection{Results}

\begin{figure}[phtb!]
  \centering
  \begin{flushleft}
   {\warp}
  \end{flushleft}
  \resizebox{5.0in}{!}{\includegraphics[page=1]{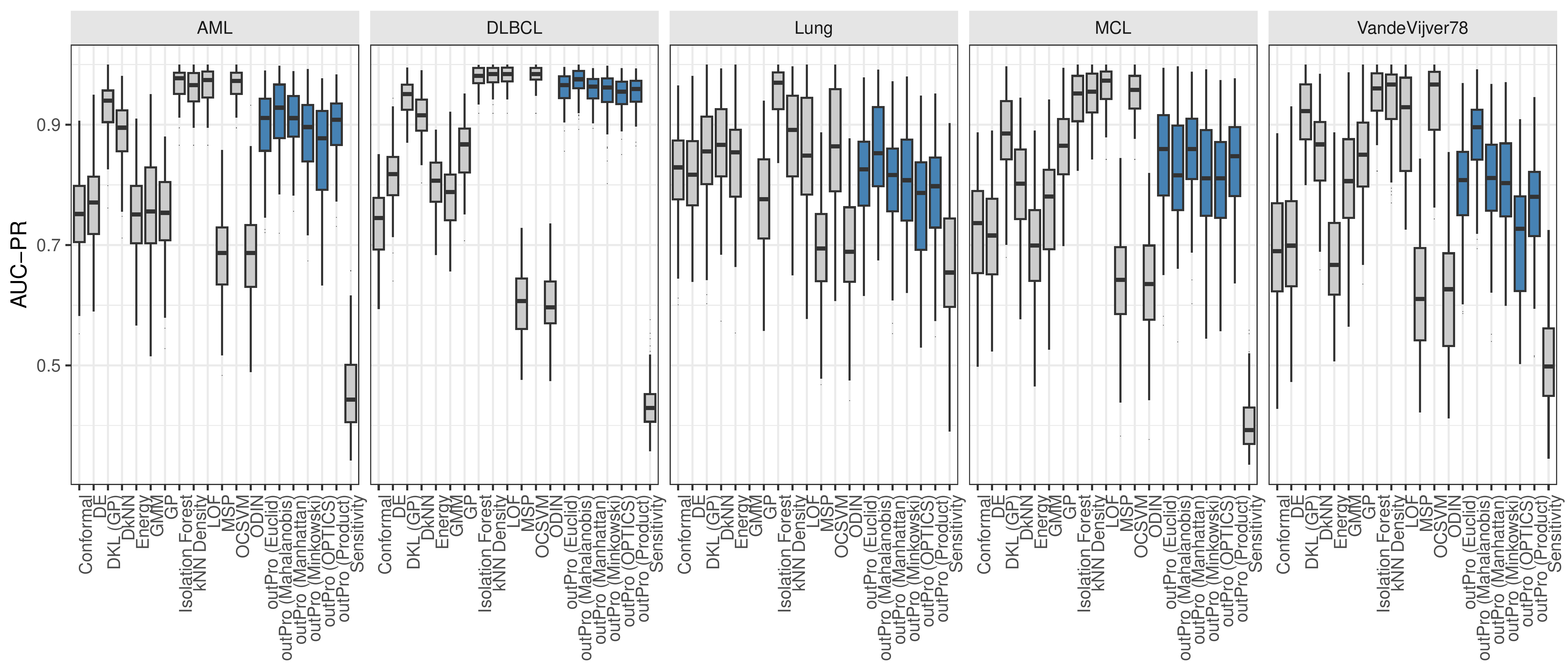}}
  \begin{flushleft}
   {\joint}
  \end{flushleft}
  \resizebox{5.0in}{!}{\includegraphics[page=2]{benchmark_realdata_highdim_box.pdf}}
  \begin{flushleft}
    {\support}
  \end{flushleft}
  \resizebox{5.0in}{!}{\includegraphics[page=3]{benchmark_realdata_highdim_box.pdf}}

  \vskip6pt

  \caption{\it AUC-PR scores for high-dimensional microarray datasets
    under the three copula-based anomaly modes, \warp, \joint, and
    \support. Each box summarizes results over 100 replications. Blue
    signifies OutPro methods.}

  \label{hd.box}
\end{figure}

\begin{figure}[phtb!]
  \centering
  %\vskip1.25in
  \resizebox{5.50in}{!}{\includegraphics[page=1]{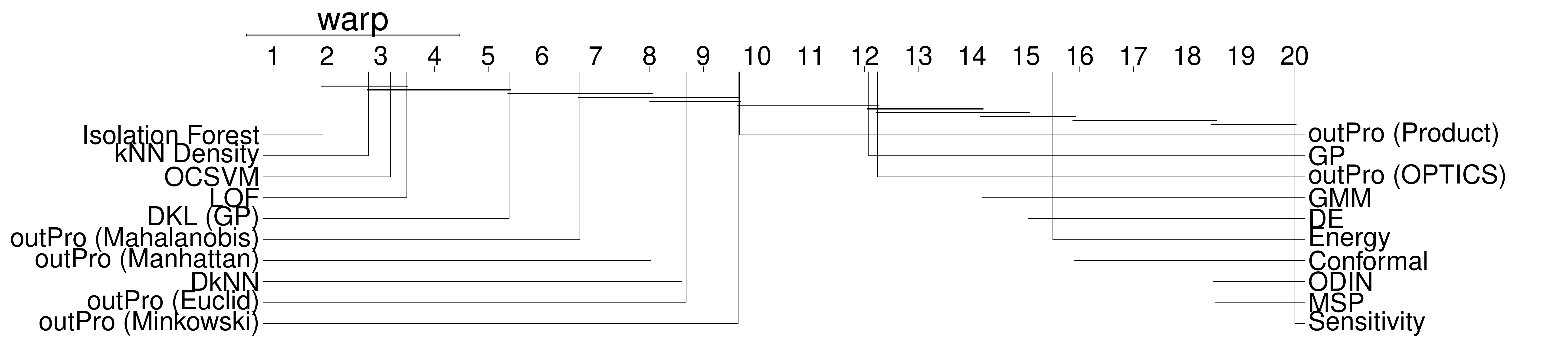}}
  \resizebox{5.50in}{!}{\includegraphics[page=2]{benchmark_realdata_highdim_cd.pdf}}
  \resizebox{5.50in}{!}{\includegraphics[page=3]{benchmark_realdata_highdim_cd.pdf}}

  \vskip10pt

  \caption{\it Critical difference (CD) plots showing average rank
    performance across the five high-dimensional microarray datasets and
    copula modes. Lower ranks indicate better performance. The \warp\
    panel is led by Isolation Forest, robust $k$NN density, OCSVM, and
    LOF. The \joint\ panel is led by OutPro distances. The \support\
    panel is more mixed, with robust $k$NN density and several OutPro
    distances near the top.}
  
   \label{hd.cd}

   \vskip20pt

  \resizebox{5.0in}{!}{\includegraphics{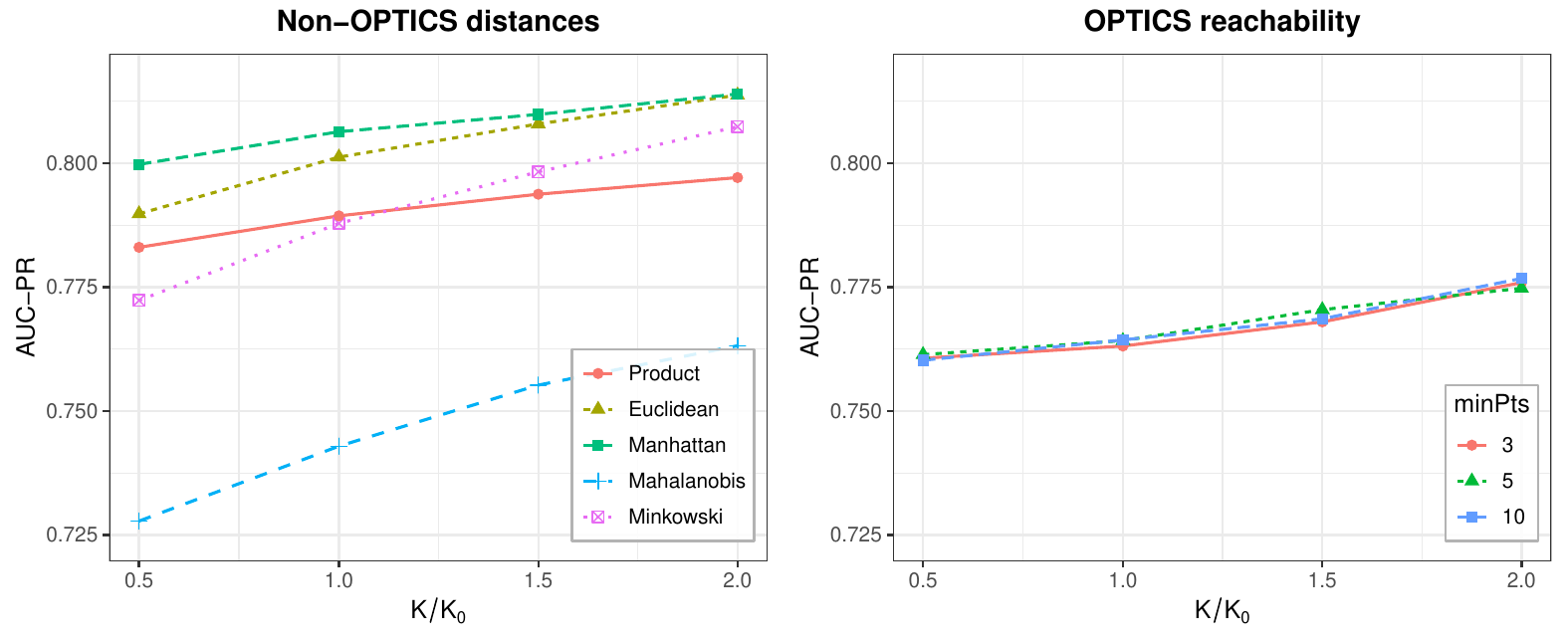}}
  \caption{\it Sensitivity of OutPro to neighborhood size $K$ under
  the \support\ anomaly mode on the 61 PMLB regression datasets.  The
  left panel shows AUC-PR for the non-OPTICS distances as a
  function of $K/K_0$.  The right panel shows
  AUC-PR for the OPTICS reachability score for different values of
  $\texttt{minPts}$.}
  \label{fig:support.sensitivity}

\end{figure}

Boxplots of AUC-PR by dataset and copula mode are shown in
\autoref{hd.box}, with an overall summary via critical difference
(CD) plots given in \autoref{hd.cd}. The ranking pattern is more mixed than
in the PMLB analysis.

Under the \warp\ mode, the best methods are Isolation Forest, robust
$k$NN density, and OCSVM, while LOF and DKL (GP) also perform
well. Among the OutPro procedures, Mahalanobis is best, while
Manhattan, Euclidean, Minkowski, and product fall somewhere in the
middle of all procedures. This again fits the structure of \warp,
which generates tail distorted points that remain largely within the
observed support and can therefore be captured well by classical
density based rules.

As in the previous PMLB study, the picture changes under the
\joint\ mode. OutPro Euclidean, Manhattan, Mahalanobis, Minkowski,
product, and OPTICS are the top procedures. The best non OutPro
methods are robust $k$NN density, GMM, DkNN, and conformal prediction.
Thus when anomalies act through dependence rather
than marginal extremes, the local predictive structure used by OutPro
is more informative than generic density or uncertainty scores.

Under the \support\ mode, the results are  mixed. Robust $k$NN
density has the best performance.  LOF is also near the top.  At
the same time, OutPro Mahalanobis, Minkowski, Euclidean, and Manhattan
occupy four of the five top spots, showing that OutPro remains highly
competitive. The product based OutPro score is less effective,
especially under \support, where it ranks below the other OutPro
methods. One explanation is that in these high-dimensional datasets
many coordinates carry signal, so a multiplicative combination of
coordinate wise discrepancies can amplify noise and dilute informative
deviations.  Additive or correlation adjusted distances such as
Manhattan, Euclidean, Minkowski, and Mahalanobis appear more stable in
this setting.

\subsection[Sensitivity Analysis for K and minPts]{Sensitivity Analysis for $K$ and \texttt{minPts}}\label{sec5.sensitivity}

To assess the stability of OutPro with respect to its main tuning
parameters, we carried out a focused ablation on the PMLB
regression datasets used in Section~\ref{sec5.pblmr}.  We chose the
\support\ copula mode for this analysis because it was the most
challenging setting in the benchmark.  Recall from
Section~\ref{sec3} that $K$ is the number of training points used for
the neighborhood $\nn_K(x)$.  By design, OutPro uses values of $K$
that are much larger than those typical of classical nearest-neighbor
methods.  The default neighborhood size used in our previous analyses
was $K_0 = \min(n/10, 5000)$; here we evaluated $K$ such that
$K/K_0 \in \{0.5, 1, 1.5, 2\}$.  For the OPTICS distance we also
varied $\texttt{minPts} \in \{3, 5, 10\}$.  Because $\texttt{minPts}$
enters only the OPTICS reachability calculation, the non-OPTICS
distances were summarized after averaging over $\texttt{minPts}$,
while OPTICS was reported separately for each value.

\subsubsection{Results}

\autoref{fig:support.sensitivity} summarizes the AUC-PR values.
Overall, we observe that every non-OPTICS distance attained its best
AUC-PR at the largest neighborhood (here equal to $2K_0$), with the
Manhattan, Euclidean and Minkowski distances achieving best overall
performance.  When using OPTICS, AUC-PR similarly improved with $K$
but was essentially flat across $\texttt{minPts} \in \{3, 5, 10\}$.

This shows OutPro is stable across its tuning parameters.  The default
value $K_0$ worked well, and while larger neighborhoods
improved performance, the gains were moderate. The monotone
improvement with $K$ is consistent with the design of the
method. Because $\nn_K(x)$ is constructed from the fitted model's rule
regions rather than from a global distance, it already concentrates on
training cases that are functionally relevant to $x$. Enlarging $K$
within this model-derived neighborhood adds useful context without
introducing the noise that would arise from expanding a conventional
nearest-neighbor search.

\subsection{Component Ablation}\label{sec5_ablation}

We next isolate the main components of OutPro using the PMLB
regression benchmark. The components are the VarPro signal set
$\widehat S$, the forest neighborhood $\nn_K(x^*)$, and the VarPro
weights used in the final distance. Here the forest neighborhood
means the set of $K$ training points with the largest OutPro proximity
scores $W(x_i)$ for the test input $x^*$, as defined in
Section~\ref{sec3_prox}.  Ordinary KNN is the set of
$K$ training points with the smallest Manhattan distance to $x^*$,
computed using the coordinates and weights specified in
Table~\ref{tab:component_ablation}.

To keep the ablation focused on algorithmic components, all settings
use the Manhattan distance and the default neighborhood size
$K_0=\min(n/10,5000)$. Table~\ref{tab:component_ablation} reports
average AUC-PR ranks across the PMLB regression benchmark. Lower
ranks indicate better performance.

\begin{table}[phtb]
\centering
\small
\renewcommand{\arraystretch}{1.15}
\caption{\it Component ablation on the PMLB regression benchmark.
Lower average AUC-PR rank is better. The column $\bar r$ is the
average rank across the three copula modes.}
\label{tab:component_ablation}
\begin{tabular}{lccc|cccc}
\hline
Setting & Space & Neighborhood & Weights
& \joint\ & \support\ & \warp\ & $\bar r$ \\
\hline
Full OutPro
& $\widehat S$ & forest & VarPro
& 2.88 & 1.40 & 1.23 & 1.84 \\

Unweighted OutPro
& $\widehat S$ & forest & uniform
& 3.38 & 3.42 & 3.80 & 3.53 \\

All-coordinate OutPro
& all & forest & VarPro
& 2.91 & 2.94 & 3.89 & 3.25 \\

Subspace KNN
& $\widehat S$ & ordinary KNN & VarPro
& 1.47 & 2.82 & 2.30 & 2.33 \\

Full-space KNN
& all & ordinary KNN & uniform
& 4.37 & 4.42 & 3.78 & 4.19 \\
\hline
\end{tabular}
\end{table}

The results support the integrated construction of OutPro.
Full OutPro has the best average rank across the three
copula modes, with especially strong performance for \support\ and \warp\
shifts. Removing VarPro weights while keeping the same selected
variables and forest neighborhood reduces performance in all three
modes. Removing the subspace restriction by using all coordinates also
reduces performance, 
consistent with signal dilution from nuisance coordinates.

The comparison with KNN shows that the effect of the forest
neighborhood depends on the type of shift. Subspace KNN has the best
rank for \joint\ shifts, but Full OutPro is substantially better for
\support\ and \warp\ shifts and is better overall. Full-space KNN performs
poorly in all settings, indicating that ordinary nearest-neighbor
search in the full covariate space is not sufficient.

\section{Evaluating Lymphadenectomy Survival Through OOD Analysis}\label{sec6}

We next consider a survival case study from the Worldwide Esophageal
Cancer Collaboration (WECC)~\citep{rice2009wecc}. The registry
includes patients who underwent esophagectomy alone for esophageal
cancer, with right censored follow-up on all-cause mortality
\citep{rizk2010optimum}. Previous WECC analyses used random survival
forests (RSF)~\citep{Ishwaran2008} to study the relation between
lymphadenectomy, measured by the number of lymph nodes resected, and
five-year survival. Those analyses found that greater lymphadenectomy
was generally associated with improved survival, with optimal counts
depending on tumor stage and histopathologic type
\citep{rizk2010optimum}. We revisit this setting to study whether
patients receiving more extensive lymphadenectomy become less
outlying relative to clinically relevant training populations.

The analysis used $n=6{,}142$ adenocarcinoma cases. Covariates
included pTNM classification, number of lymph nodes resected, number
of positive nodes, histologic grade, tumor location and length,
residual cancer status, and patient demographics, giving
$d=35$ variables. In the pTNM system, the prefix ``p'' denotes
pathologic staging from surgical resection specimens, and the T
category records depth of tumor invasion from pT1 to pT4. We focus on
patients with pT3 or pT4 tumors, combined into a single group.

\begin{figure}[phtb!]
  \centering
  \resizebox{4.50in}{!}{\includegraphics[page=1]{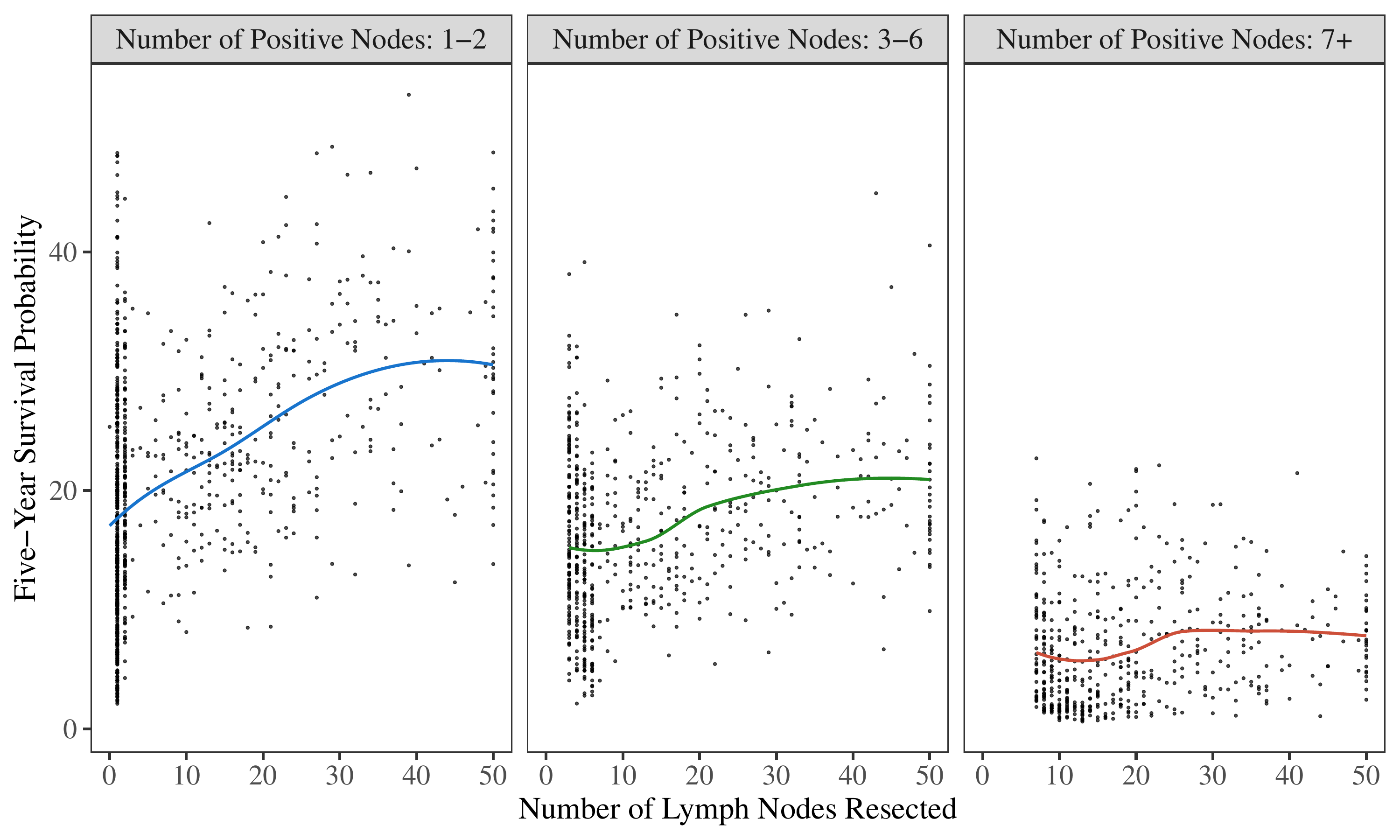}}

  \vskip25pt

  \resizebox{4.0in}{!}{\includegraphics[page=2]{wecc.pdf}}

  \caption{\it WECC adenocarcinoma analysis for pT3--4 tumors.
    Upper panel shows out-of-bag five-year survival predictions from
    RSF as a function of the number of lymph nodes removed, stratified
    by the number of positive nodes. Survival generally improves with
    greater lymphadenectomy, with smaller gains as nodal involvement
    increases. Lower panel shows mean OOD percentile score as a
    function of the lymphadenectomy cutoff, averaged over 31 hold-out
    scenarios with cutoffs from 0 to 30. A score of 100 corresponds to
    the most extreme OOD score under the null training distribution.}

  \label{wecc}
\end{figure}

The upper panel of \autoref{wecc} gives the RSF out-of-bag five-year
survival predictions for pT3--4 adenocarcinoma, stratified by the
number of positive nodes, 1--2, 3--6, and $\geq 7$. The pattern
recovers the clinical signal reported previously: survival improves
as more lymph nodes are removed, but the gain is attenuated in
patients with heavier nodal involvement.

For OOD scoring, we created 31 train/test scenarios by varying the
lymphadenectomy cutoff from 0 to 30 nodes. At each cutoff, the test
set contained node-positive pT3--4 patients whose lymph node count met
or exceeded the cutoff, and all remaining patients formed the training
set. A cutoff of 0 therefore holds out all node-positive pT3--4 cases,
a clinically distinct group with deeply invasive disease and nodal
involvement. As the cutoff increases, the held-out group is restricted
to patients with progressively greater lymphadenectomy. For each
scenario, we computed the mean OOD percentile score for the held-out
group on a 0--100 scale.

The lower panel of \autoref{wecc} shows that mean OOD scores decrease
as the cutoff increases, indicating that patients with greater
lymphadenectomy become less distinct from the corresponding training
population. The decline is steepest for patients with 1--2 positive
nodes and shallowest for those with $\geq 7$ positive nodes. This is
consistent with the survival curves in the upper panel, where the
benefit of additional lymphadenectomy is less pronounced with heavier
nodal involvement. Using the 95th percentile as a threshold for
considering a case non-OOD, the required lymphadenectomy count is
approximately 3 nodes for the 1--2 and 3--6 positive-node categories,
but about 30 nodes for the $\geq 7$ category. At a stricter 90th
percentile threshold, the required count rises to 5 and 10 nodes for
the 1--2 and 3--6 categories, respectively.

Because the surgeon does not know the true nodal status during
resection, the OOD analysis supports a more aggressive
lymphadenectomy for suspected deeply invasive tumors, on the order of
30 nodes. This finding is consistent with the earlier WECC report,
which estimated optimal lymphadenectomy counts for pT3/T4 cancers
ranging from 29 to 50 nodes depending on histopathologic type
\citep{rizk2010optimum}.

\section{Summary and Limitations}\label{sec7}

We have proposed an integrated model-aware, subspace-aware framework
for OOD detection in supervised settings. A key feature of the method
is that the OOD score is not constructed from the model's predicted
values, residuals, or uncertainty measures, but rather from the rule
structure learned by a supervised forest. Those learned rules identify
a reference neighborhood for each test input and select a predictive
subspace, after which OOD is measured by distances computed only within
that subspace. The fitted model therefore enters the detector through
its learned partition of the covariate space rather than through its
output, which focuses the score on covariate variation that is relevant
for the outcome and separates functionally meaningful departures from
shifts occurring mainly in irrelevant coordinates. The framework is
general, and the empirical work in this paper focused on regression and
survival settings.

The empirical results show that performance depends on the form of the
shift. In the Friedman experiments, OutPro performed best for small
feature-targeted perturbations, where anomalies are subtle and
difficult to distinguish from ordinary test variation. In the PMLB
benchmarks, OutPro was best under the \joint\ and
\support\ modes. Under the \warp\ mode, however, classical tabular
methods such as Isolation Forest, one-class SVM, Local Outlier Factor,
and robust $k$NN density were often among the best procedures. In the
high-dimensional microarray studies, OutPro again performed very well
under \joint\ and remained competitive under \support, while the
\warp\ results were more mixed. This pattern is consistent with the
structure of \warp, which creates low-density marginal tail distortion
within the observed support and is therefore well matched to density
and isolation rules. A sensitivity analysis also suggested that the
method is reasonably stable over a broad range of its key tuning
parameters, including $K$ (the number of neighbors), with only moderate
gains from enlarging the model-derived neighborhood.

The case study on esophageal cancer and lymphadenectomy illustrated the
clinical interpretability of the resulting scores. Using pT3-4
adenocarcinoma patients from the WECC dataset, we examined how
increasingly aggressive lymphadenectomy moved patients toward or away
from anomalous regions relative to the training cohort. The analysis
suggested that for suspected deeply invasive tumors, removal of
approximately 30 lymph nodes may be a reasonable surgical target when
true nodal status is unknown, in line with prior WECC findings.

OutPro was also computationally manageable across all studies
considered. Runtimes across the PMLB regression benchmarks and the
high-dimensional microarray datasets were well under 30 seconds at
their largest, and in the microarray studies were especially short
because variable prioritization substantially reduced the working
dimension and sample sizes were moderate. The computation benefits from
three features of the procedure, namely that the prediction engine is
based on random forests, that priority rules can be extracted
efficiently from the trained ensemble, and that the final neighborhood
scoring step is inexpensive once the predictive subspace has been
selected.

Several limitations should be mentioned. First, the empirical results
show that no single detector is uniformly best, and within OutPro no
single subspace distance dominates across all settings. The product
score worked well in lower-dimensional settings, while Manhattan,
Euclidean, Minkowski, and Mahalanobis scores were often more stable in
higher-dimensional problems, and classical tabular methods were
especially competitive under \warp. A fuller theory for choosing the
subspace distance and identifying the settings where each score is
preferred remains open. Second, the method depends on estimating a
predictive subspace from the fitted model, and errors in that step can
affect both the reference neighborhood and the final score. Third,
while the copula-based benchmarks provide varied and controlled
departures, any synthetic benchmark remains an approximation to the
shifts encountered in practice.
A fourth limitation concerns pure concept shift. Although the trained
forest, signal variables, and reference neighborhoods are learned from
labeled training data, the OutPro score for a new input is computed
from its covariates. Consequently, two test distributions with the same
$\PP^*_X$ but different $\PP^*_{Y\mid X}$ induce the same distribution
of OutPro scores. The method is therefore intended for prediction
relevant covariate departures, not for detecting conditional shifts
that leave the covariate distribution unchanged.

%\vskip25pt
%\noindent{\bf Data and Code Availability.}
\subsection*{Data and Code Availability}
Our code is publicly available as a CRAN R-package \varPro.  All
simulated datasets can be reproduced directly from the code.  Public
benchmark datasets used in this study are available from the PMLB
repository, and information on the microarray datasets can be found in
the cited references.  The esophageal cancer data from the Worldwide
Esophageal Cancer Collaboration are not publicly available.

%\vskip10pt\noindent{\bf Acknowledgements.}
\subsection*{Acknowledgements}
This work was supported by the National Institute of General Medical
Sciences of the National Institutes of Health under Award Number R35
GM139659 and by the National Heart, Lung, and Blood Institute of the
National Institutes of Health under Award Number R01 HL164405.

%\vskip10pt\noindent{\bf Conflict of Interest.}
\subsection*{Conflict of Interest}
The authors declare no conflicts of interest.

\subsection*{CRediT Authorship Contribution Statement}
{\bf Min Lu:} Conceptualization, Methodology, Supervision.
{\bf Hemant Ishwaran:} Conceptualization, Methodology, Writing -- Original Draft,
Software, Supervision, Writing -- Review \& Editing, Funding
Acquisition.

\vskip25pt

%%%%%%%%%%%%%%%%%%%%%%%%%%%%%%%%%%%%%%%%%%%%%%%%%%%%%%%%%%%%%%%%%%%%%%%%%%%%%%%%%%%%
%%%
%%%
%%% bibliography
%%%
%%%
%%%%%%%%%%%%%%%%%%%%%%%%%%%%%%%%%%%%%%%%%%%%%%%%%%%%%%%%%%%%%%%%%%%%%%%%%%%%%%%%%%%%

{
  \catcode`'=9
  \catcode``=9

  \setlength{\bibsep}{4.5pt}  % Adjust space between references

  \bibliographystyle{unsrtnat-init}

  \bibliography{references}

}

\end{document}